%% file: iclr2026_conference.tex
\documentclass{article} 
\usepackage{iclr2026_conference,times}

\input{math_commands.tex}

\usepackage{hyperref}
\usepackage{url}
\usepackage{natbib}
\usepackage{amsmath}
\usepackage{caption}
\usepackage{booktabs}
\usepackage[table]{xcolor}
\usepackage{multirow}
\usepackage{graphicx}
\usepackage{float}
\usepackage{pifont}
\usepackage{amsfonts}
\usepackage{nicefrac}
\usepackage{microtype}
\usepackage{amssymb, amsthm}
\usepackage{colortbl}
\usepackage{multicol}
\usepackage{makecell}
\usepackage{diagbox}
\usepackage{subcaption}
\usepackage[utf8]{inputenc}
\usepackage{svg}
\usepackage{comment}
\usepackage{enumitem}

\title{\vspace{-8mm}Let Features Decide Their Own Solvers:\\Hybrid Feature Caching for Diffusion\\Transformers\vspace{-1mm}}

\author{
\textbf{Shikang Zheng}$^{1,2}$\quad
\textbf{Guantao Chen}$^{1}$\quad
\textbf{Qinming Zhou}$^{1,3}$\quad
\textbf{Yuqi Lin}$^{1}$\quad
\textbf{Lixuan He}$^{1,3}$\\[1.1pt]
\hspace{0.23em}\textbf{Chang Zou}$^{1}$\quad
\textbf{Peiliang Cai}$^{1}$\quad
\textbf{Jiacheng Liu}$^{1}$\quad
\textbf{Linfeng Zhang}$^{1\dag}$\\[1.6pt]
$^{1}$Shanghai Jiao Tong University\hspace{0.4em}
$^{2}$South China University of Technology\hspace{0.4em}
$^{3}$Tsinghua University
}

\iclrfinalcopy 
\begin{document}

\maketitle

\renewcommand{\thefootnote}{\fnsymbol{footnote}}
\footnotetext[2]{Corresponding author.}
\renewcommand{\thefootnote}{\arabic{footnote}}

\begin{abstract}
\vspace{-2mm}
Diffusion Transformers offer state-of-the-art fidelity in image and video synthesis, but their iterative sampling process remains a major bottleneck due to the high cost of transformer forward passes at each timestep. To mitigate this, feature caching has emerged as a training-free acceleration technique that reuses or forecasts hidden representations. However, existing methods often apply a uniform caching strategy across all feature dimensions, ignoring their heterogeneous dynamic behaviors. Therefore, we adopt a new perspective by modeling hidden feature evolution as a mixture of ODEs across dimensions, and introduce \textbf{HyCa}, a Hybrid ODE solver inspired caching framework that applies dimension-wise caching strategies. HyCa achieves near-lossless acceleration across diverse domains and models, including 5.55$\times$ speedup on FLUX, 5.56$\times$ speedup on HunyuanVideo, 6.24$\times$ speedup on Qwen-Image and Qwen-Image-Edit without retraining.
\end{abstract}

\input{sec/2_intro}

\input{sec/3_related_work}
\input{sec/4_method}

\input{sec/5_exp}

\input{sec/6_discussion}
\input{sec/7_conclusion}

\bibliography{iclr2026_conference}
\bibliographystyle{iclr2026_conference}

\end{document}

%% file: math_commands.tex
\usepackage{amsmath,amsfonts,bm}

\def\eqref#1{equation~\ref{#1}}

\def\1{\bm{1}}

\DeclareMathAlphabet{\mathsfit}{\encodingdefault}{\sfdefault}{m}{sl}
\SetMathAlphabet{\mathsfit}{bold}{\encodingdefault}{\sfdefault}{bx}{n}



%% file: sec/2_intro.tex
\section{Introduction}

\vspace{-3mm}
\begin{figure}[htp]
  \centering
  \includegraphics[trim=0 0 0 0, clip,width=1\linewidth]{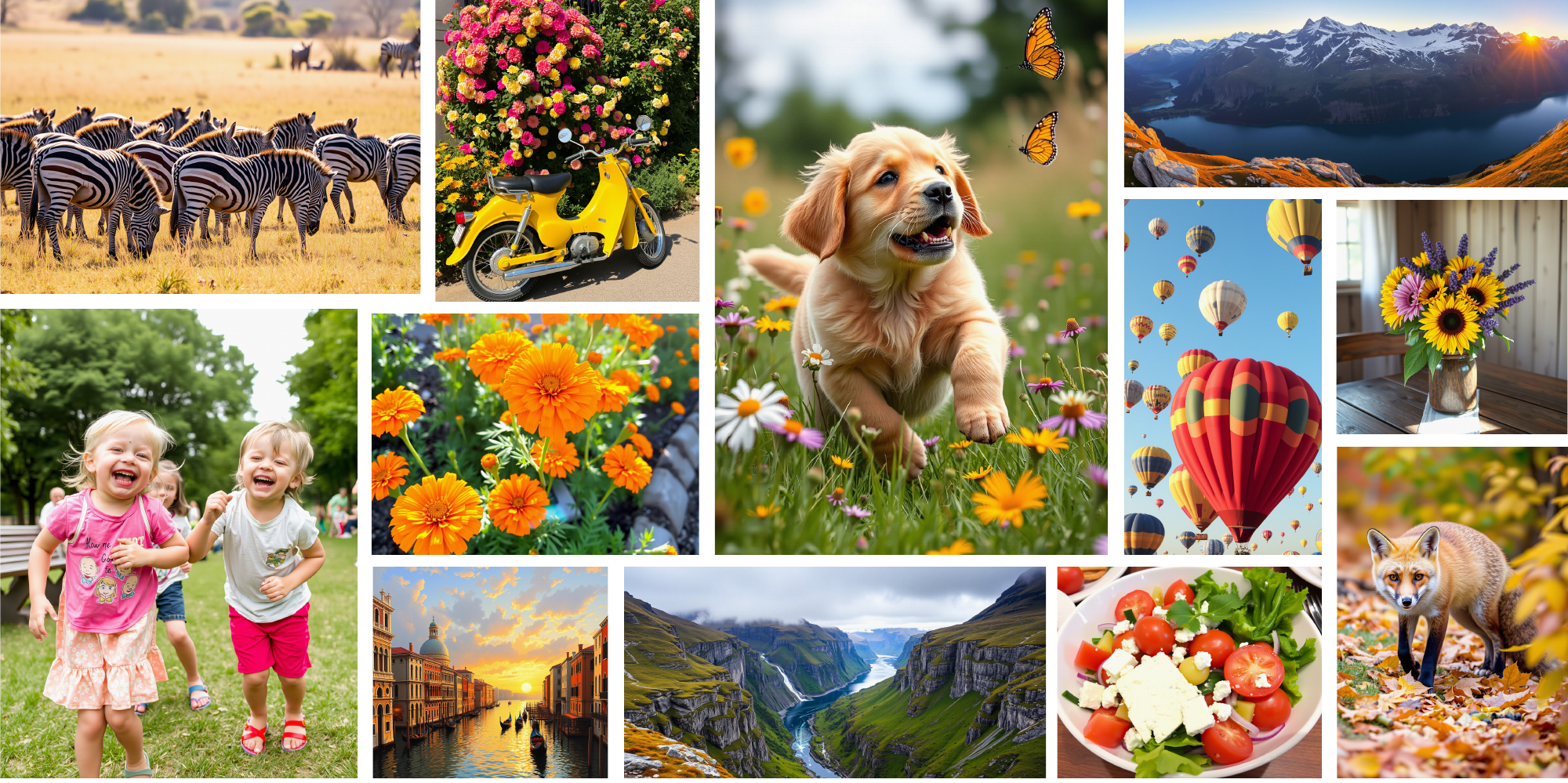}
  \vspace{-5mm}
  \caption{Images generated on Qwen-Image with HyCa at \textbf{6.24$\times$} acceleration.}
  \vspace{-2mm}
  \label{fig:poster}
\end{figure}

Diffusion Transformers (DiTs) have recently achieved impressive success across image and video generation tasks, demonstrating strong modeling capacity and generation quality. However, the iterative nature of diffusion sampling presents a significant bottleneck, as each output demands multiple transformer passes. This high computational cost hinders deployment in scenarios with strict latency or resource constraints, driving the ongoing research on efficient inference methods.

\begin{figure}[htbp]
  \centering
  \includegraphics[trim=180 55 180 90, clip,width=1\linewidth]{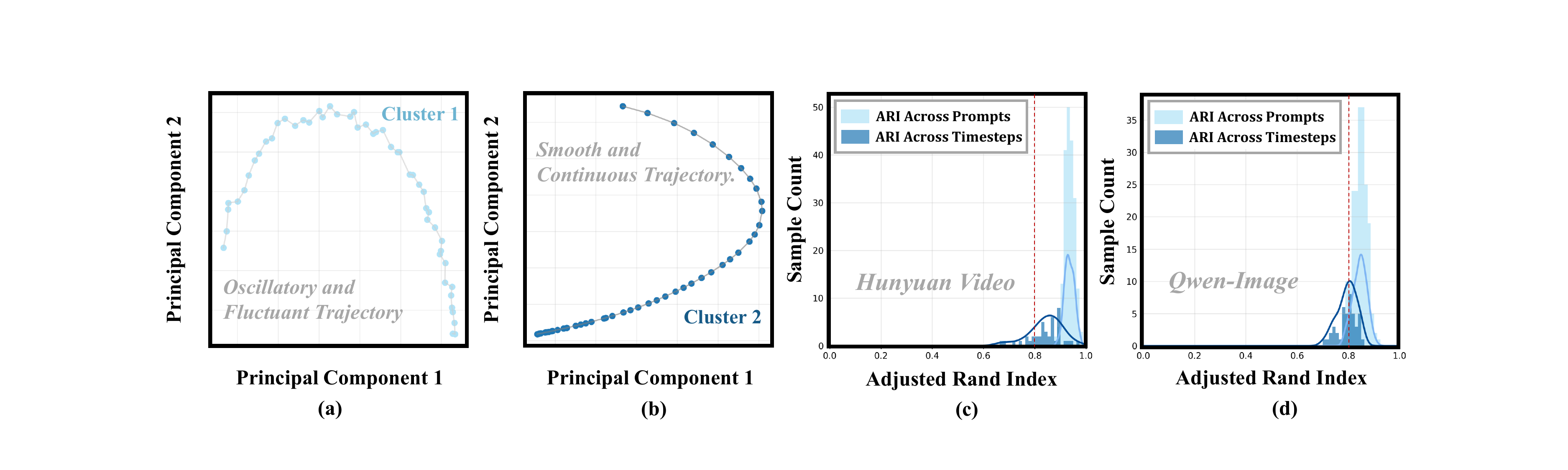}
  \caption{\textbf{Feature trajectory clusters and stability of assignments.} (a--b) Cluster~1 shows oscillatory trajectories while Cluster~2 shows smooth ones. (c--d) ARI distributions on Hunyuan Video and Qwen-Image exceed 0.8 in most cases, confirming stable and consistent cluster assignments across prompts and timesteps. An ARI above 0.8 indicates strong agreement and high clustering reliability.}
 \vspace{-5mm}
  \label{fig:intro}
\end{figure}

To address this challenge, two primary acceleration directions have emerged: reducing the total number of sampling steps via algorithmic advancements~\citep{lu2022dpm}, and lowering the cost of each step through architectural optimization~\citep{yuan2024ditfastattnattentioncompressiondiffusion, zhao2025realtimevideogenerationpyramid}. Among these, training-free feature caching has emerged as a promising solution. It exploits the temporal coherence of hidden representations by reusing features, thereby reducing redundant computation. Early works such as DeepCache~\citep{ma2024deepcache} demonstrated the feasibility of this idea in U-Net backbones, recent methods such as FORA~\citep{selvaraju2024fora}, ToCa~\citep{zou2024accelerating}, TaylorSeer~\citep{liu2025reusingforecastingacceleratingdiffusion} extended caching to transformer-based architectures and showed that feature caching can be effectively viewed as solving the temporal evolution of hidden features. Despite progress, current approaches are still limited in critical ways.



Existing methods implicitly assume that all hidden dimensions evolve under a single, unified system. However, this assumption is untenable in DiTs, where the feature space is high-dimensional and exhibits complicated behaviors. Such complexity is unlikely to be captured by a single process. To further investigate, we analyze how each feature dimension changes over timesteps and group them into clusters based on their dynamics. As shown in Fig.~\ref{fig:intro}(a), some dimensions fluctuate sharply with oscillatory patterns, indicating stiffness or multimodal behavior, while others evolve smoothly and predictably, reflecting stable dynamics on Fig.~\ref{fig:intro}(b). These observations suggest that the feature space of DiTs is better described as a complex system, where different groups of dimensions follow distinct temporal patterns, highlighting the need for tailored solvers rather than a one-size-fits-all approach.

Therefore, we introduce \textbf{HyCa}, a hybrid caching framework that models hidden feature evolution as a mixture of ODEs and applies suitable solvers for every dimension. Hyca begins with unsupervised clustering, grouping dimensions with similar dynamic behaviors, and modeling them into a shared ODE. Then, HyCa assigns the most suitable solver to each cluster. Normally, identifying the best solver would require running inference on a large set of images and comparing quantitative metrics. However, surprisingly, we found that cluster assignments are highly stable across resolutions, timesteps, and even prompts. As shown in Fig.~\ref{fig:intro}(c)(d), this invariance allows us to evaluate solver performance on a \textbf{single prompt} at a \textbf{single timestep} to reliably identify the best solver, achieving results comparable to large-scale evaluation. Thus, with \emph{``One-Time Choosing''} performed offline for each model, \emph{``All-Time Solving''} becomes possible without any additional cost during inference.

HyCa provides robust and adaptive feature prediction across diverse tasks and architectures. Without retraining, it achieves near-lossless acceleration of \textbf{5.56$\times$} on FLUX and Hunyuan Video, \textbf{6.24$\times$} on Qwen-Image and Qwen-Image-Edit. Moreover, it is also fully compatible with distillation, reaching up to \textbf{24.4$\times$} speedup on FLUX and \textbf{12.2$\times$} on Qwen-Image while maintaining strong image quality. In summary, our main contributions are:

\begin{itemize}[leftmargin=10pt,topsep=0pt]
\item \textbf{Heterogeneous Feature Dynamics.} We show that feature dimensions in DiTs do not follow a single unified system but exhibit heterogeneous dynamic behaviors that are better described as a mixture of ODEs. Through dynamics clustering analysis across multiple settings, we further reveal that these cluster's distributions are consistent and input-invariant.

\item \textbf{HyCa Framework.} Inspired by hybrid ODE solvers in numerical analysis, we propose HyCa, a training-free framework that groups feature dimensions by their dynamics and automatically assigns the most suitable solver to each group, with minimal overhead.

\item \textbf{Outstanding Performance.} We evaluate HyCa across diverse architectures and tasks, including Drawbench on FLUX and Qwen-Image, Vbench on HunyuanVideo, GEdit-Bench on Qwen-Image-Edit, and even distilled models. In all settings, HyCa delivers state-of-the-art performance.
\end{itemize}

%% file: sec/3_related_work.tex
\section{Related Work}\label{sec:related-work}

Diffusion models~\citep{sohl2015deep,ho2020DDPM} have achieved strong image/video generation quality. Early U-Net backbones~\citep{ronneberger2015unet} faced scaling limits that Diffusion Transformers (DiT)~\citep{peebles2023dit} alleviated, enabling rapid progress across modalities and resolutions~\citep{chen2023pixartalpha,chen2024pixartsigma,opensora,yang2025cogvideox}. Nevertheless, the iterative nature of sampling remains a key inference bottleneck. Two complementary research lines thus emerge: (i) reducing the number of steps and (ii) reducing the cost per step. Beyond speed, a central challenge is maintaining stability and fidelity under aggressive acceleration, especially when feature dynamics are heterogeneous across dimensions and timesteps.

\subsection{Sampling Timestep Reduction}
DDIM~\citep{songDDIM} introduced deterministic few-step sampling that preserves perceptual quality. Higher-order ODE solvers (DPM-Solver and variants)~\citep{lu2022dpm,lu2022dpm++,zheng2023dpmsolvervF} improve accuracy–cost trade-offs via multi-step/multi-stage discretizations with carefully controlled local truncation error. Rectified Flow~\citep{refitiedflow} shortens transport paths, while distillation~\citep{salimans2022progressive} compresses long trajectories into compact generators. Consistency models~\citep{song2023consistency} enable few-step synthesis by learning a direct noise-to-clean mapping.

\subsection{Denoising Network Acceleration}
\noindent\textbf{Model Compression Acceleration.} Pruning~\citep{structural_pruning_diffusion,zhu2024dipgo}, quantization~\citep{10377259,shang2023post,kim2025ditto}, distillation~\citep{li2024snapfusion}, and token reduction~\citep{bolya2023tomesd,kim2024tofu,zhang2024tokenpruningcachingbetter,zhang2025sito,cheng2025catpruningclusterawaretoken} reduce compute with limited runtime overhead. While effective, they typically require additional training and may degrade robustness under domain shifts if the compression is too aggressive.

\noindent\textbf{Feature Caching Acceleration.} Feature caching reuses activations to avoid redundant computation. Early U-Net methods~\citep{li2023FasterDiffusion,ma2024deepcache} inspired DiT-specific designs: FasterCache~\citep{lv2025fastercachetrainingfreevideodiffusion}, FORA~\citep{selvaraju2024fora}, $\Delta$-DiT~\citep{chen2024delta-dit}, TeaCache~\citep{liu2024timestep}, and DiTFastAttn~\citep{yuan2024ditfastattn}. Dynamic updates (ToCa/DuCa)~\citep{zou2024accelerating,zou2024DuCa}, unified cache–prune pipelines~\citep{sun2025unicpunifiedcachingpruning}, and region-adaptive sampling~\citep{liu2025regionadaptivesamplingdiffusiontransformers} further improve efficiency. Among these advances, TaylorSeer~\citep{liu2025reusingforecastingacceleratingdiffusion} exemplifies the \emph{cache-then-forecast} paradigm by polynomial extrapolation from cached neighbors.

%% file: sec/4_method.tex
\section{Method}

\subsection{Preliminary}

\noindent\textbf{Diffusion Models.} Diffusion models~\citep{ho2020DDPM, songDDIM} generate structured data by progressively refining random noise through a series of denoising steps. At each timestep \(t\), the model predicts a conditional Gaussian distribution over \(x_{t-1}\) given \(x_t\), where both the mean and variance are parameterized. This generative process can be formulated as:
\begin{equation}
p_\theta(x_{t-1} | x_t) = \mathcal{N} \left( x_{t-1}; \frac{1}{\sqrt{\alpha_t}} \left( x_t - \frac{1 - \alpha_t}{\sqrt{1 - \bar{\alpha}_t}} \tau_\theta(x_t, t) \right), \beta_t \mathbf{I} \right),
\end{equation}
where \(\mathcal{N}\) denotes a normal distribution, \(\alpha_t\) and \(\beta_t\) are noise schedule parameters, and \(\tau_\theta(x_t, t)\) denotes the model's estimate of the noise component. Sampling begins from a pure noise vector and proceeds by repeatedly drawing samples from these intermediate distributions until a clean image is produced.

\noindent\textbf{Diffusion Transformer Architecture.} The Diffusion Transformer (DiT)~\citep{DiT} adopts a hierarchical design, expressed as a composition of modules \(\mathcal{G} = g_1 \circ g_2 \circ \cdots \circ g_L\). Each module \(g_l\) consists of a self-attention layer (\(\mathcal{F}_{\text{SA}}^l\)), a cross-attention layer (\(\mathcal{F}_{\text{CA}}^l\)), and a feedforward MLP (\(\mathcal{F}_{\text{MLP}}^l\)). These components are dynamically modulated across timesteps to accommodate the evolving noise levels during generation. The input \(\mathbf{x}_t = \{x_i\}_{i=1}^{H \times W}\) is represented as a sequence of patch tokens. Each module includes a residual update of the form \(\mathcal{F}(\mathbf{x}) = \mathbf{x} + \text{AdaLN} \circ f(\mathbf{x})\), where AdaLN (adaptive layer normalization) conditions the normalization parameters on the noise timestep, allowing for more effective denoising across varying noise scales.

\begin{figure}[htbp]
  \centering
  \includegraphics[trim=35 85 305 15, clip,width=1\linewidth]{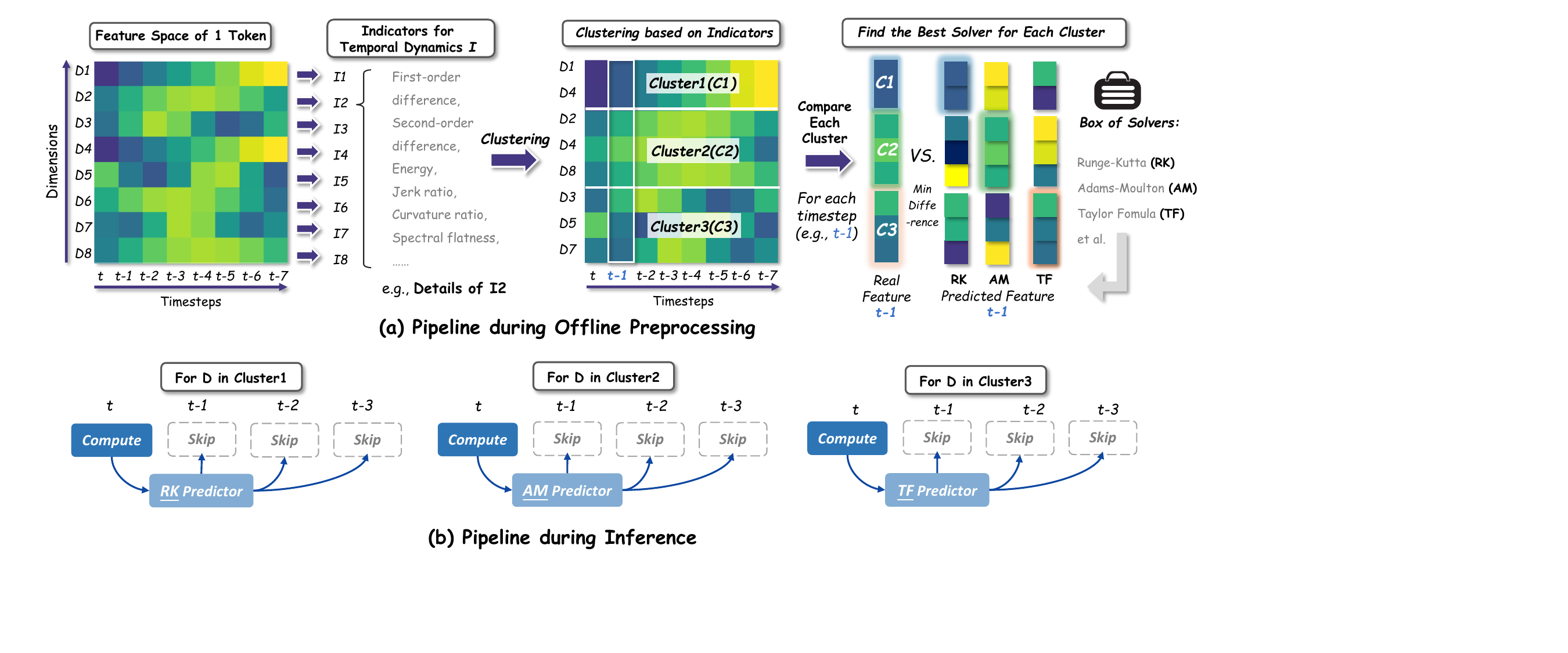}
  \caption{\textbf{HyCa Framework.} (a) Offline Preprocessing: feature dimensions are first analyzed and clustered with temporal indicators (e.g., differences, curvature). For each cluster, candidate solvers generate predicted features, then compared against real computed features; the solver with minimum error is then assigned to that cluster. (b) Inference: once assigned, each cluster consistently reuses its solver, enabling efficient prediction by skipping redundant computations while maintaining accuracy.}

  \label{fig:model}
\end{figure}

\noindent\textbf{Feature Caching.} Feature caching aims to reduce the cost of diffusion sampling by avoiding repeated computation of hidden features across timesteps. At each timestep \( t \), the model produces hidden features \(\mathcal{F}_t = \{ \mathcal{F}_t^l \}_{l=1}^{L}\), and a caching function \(\mathcal{C}(\mathcal{F}_A, k)\) estimates features \(\tilde{\mathcal{F}}_k\) at a future timestep \(k\) using cached features from earlier timesteps. A common strategy is to reuse features from the last computed step:
\begin{equation}
\tilde{\mathcal{F}}_k = \mathcal{C}(\mathcal{F}_t, k) := \mathcal{F}_t, \quad \forall k \in (t, t + n - 1],
\end{equation}

which provides up to \((n{-}1)\times\) speedup. Recent methods improve reuse by forecasting future features, yet their reliance on a uniform prediction strategy across all dimensions often proves unstable in DiT's complex hidden feature space. In this work, we propose a hybrid approach that assigns suitable solvers for every dimension according to their dynamic behaviors.

\subsection{Feature Caching as Hybrid ODE Solving}

During reverse-time denoising in diffusion models, the hidden features evolve across timesteps. Let \(x_t\) be the latent variable at time \(t\), and let \(\mathcal{F}(x_t)\) denote the hidden feature extracted from it. Since the generative model is differentiable and \(x_t\) follows a continuous reverse-time trajectory, the composite feature map \(t \mapsto \mathcal{F}(x_t)\) is also differentiable. By the chain rule and the probability flow ODE governing \(x_t\), the feature dynamics satisfy:
\begin{equation}
  \frac{d}{dt}\,\mathcal{F}(x_t) = g_\theta\bigl(\mathcal{F}(x_t),\, t\bigr),
  \label{eq:feature_ode}
\end{equation}
where \(g_\theta\) captures the implicit time-dependent vector field induced by the underlying network weights and structure. Although \(g_\theta\) is not directly accessible in closed-form, we can sample the trajectory \(\{ \mathcal{F}(x_{t_k}) \}\) on a discrete timestep grid, enabling numerical integration using only cached feature values. This perspective naturally casts feature caching as a numerical ODE solving problem. Instead of performing full forward computation at every timestep, we aim to solve the next feature value using prior ones:
\[
\hat{\mathcal{F}}_{t+1} \approx \text{Solver}(\mathcal{F}_t, \mathcal{F}_{t-1}, \dots)
\]
To accommodate diverse local feature dynamics, from smooth near-linear segments to rapidly varying regions, we adopt a hybrid solver strategy. Concretely, we define a predictor pool \(\mathcal{S}\) that includes both explicit and implicit numerical solvers with different stability and accuracy properties, including Runge-Kutta(RK), Adams-Bashforth(AB), Taylor Fomula(TF), Backward Differentiation Formula(BDF) and Adams-Moulton(AM). This diverse solver set enables HyCa to assign methods tailored to local feature dynamics. \emph{Please refer to A.2.2 for detailed mathematical formulations.}

\subsection{HyCa Framework}

Building on this foundation, \textbf{HyCa} is designed as a feature caching framework that models hidden dynamics as a mixture of ODEs and automatically assigns the most suitable solver to each cluster through a one-time optimization procedure. HyCa begins by analyzing the temporal dynamic behavior of each feature dimension. During a probe pass on a single prompt at the first few timesteps, we extract a descriptor vector \( \phi_d \in \mathbb{R}^k \) for each feature dimension \( d \in \{1, \dots, D\} \), capturing dynamic indicators such as Jerk ratio and curvature ratio. Then, we apply \(k\)-means clustering to obtain a partition \( \{c(d)\} \), where each dimension \( d \) is assigned to a cluster \( c \in \{1, \dots, C\} \). These cluster assignments remain stable across prompts, timesteps and resolutions, thus reused throughout inference. 

The resulting clusters represent groups of dimensions that share similar temporal behaviors, enabling solver assignments to be conducted independently for each cluster. Given a solver pool \( \mathcal{S} \), HyCa selects the optimal solver \( s_c^\star \in \mathcal{S} \) for each cluster \( c \) by minimizing the average next-step prediction error across all dimensions in that cluster:
\begin{equation}
\min_{\{s_c \in \mathcal{S}\}_{c=1}^C} \sum_{c=1}^C \left[ \frac{1}{|c|} \sum_{d \in c} \left\| \hat{\mathcal{F}}_{t+1}^{(s_c, d)} - \mathcal{F}_{t+1}^{(d)} \right\|_2^2 \right],
\end{equation}
where \( \hat{\mathcal{F}}_{t+1}^{(s_c, d)} \) denotes the predicted feature for dimension \( d \) at timestep \( t+1 \) using solver \( s_c \). This formulation enables per-cluster solver selection via a one-time probing pass, ensuring that HyCa combines efficiency with the adaptability of hybrid solvers.

%% file: sec/5_exp.tex
\section{Experiments}

\subsection{Experiment Settings}

\noindent\textbf{Model Configurations.} We conduct experiments on four representative diffusion-based models: the text-to-image models {FLUX.1-dev}~\citep{flux2024} and {Qwen-Image}~\citep{liu2023rectified}, the text-to-video model {HunyuanVideo}~\citep{sun_hunyuan-large_2024}, and the image editing model {Qwen-Image-Edit}~\citep{liu2023rectified}. To further assess compatibility with model compression techniques, we also evaluate our method on distilled models: {FLUX.1-schnell} and {Qwen-Image-Lightning}. All models are evaluated under official or recommended configurations on standard public checkpoints.

\noindent\textbf{Evaluation and Metrics.} For text-to-image generation, we follow the DrawBench~\citep{saharia2022drawbench} protocol and evaluate all models on a fixed set of 200 prompts. We evaluate images using ImageReward~\citep{xu2023imagereward} for photorealism, CLIP Score~\citep{hessel2021clipscore} for text–image alignment, and PSNR, SSIM, LPIPS for fidelity. For text-to-video generation, we evaluate HunyuanVideo on VBench~\citep{VBench}, which provides multi-dimensional human-aligned assessments of motion quality, visual appearance, and semantic consistency. For image editing tasks, we use GEdit-Bench~\citep{Gedit} to evaluate model performance across a diverse set of edit types and prompts. Unless otherwise specified, all evaluations are conducted using fixed random seeds and default inference settings. \emph{Additional implementation details please refer to A.1.}

\noindent
\input{tab/qwenimage}

\noindent
\input{tab/flux_full}

\subsection{Results on Text-to-Image Generation}

As shown in Table~\ref{table:Qwenimage} HyCa achieves the best overall trade-off on \textbf{Qwen-Image} across all compression levels. At $\mathcal{N}{=}3$, it matches TaylorSeer in speed (2.78$\times$) but yields higher quality (ImageReward \textbf{1.2363} vs. 1.0685, and highest PSNR \textbf{30.42}). At $\mathcal{N}{=}6$, HyCa remains strong (\textbf{1.1939}, \textbf{29.65}), outperforming TaylorSeer (0.9483), ToCa (0.9673), and FORA (0.7767). Even at $\mathcal{N}{=}8$, it sustains good visual quality (\textbf{1.0811} at 6.25$\times$), while others drop sharply (ToCa 0.6326, FORA 0.4781). These results highlight HyCa’s robustness under high acceleration while preserving visual fidelity.

\begin{figure}[htbp]
  \centering
  \includegraphics[trim=20 10 20 10, clip,width=1\linewidth]{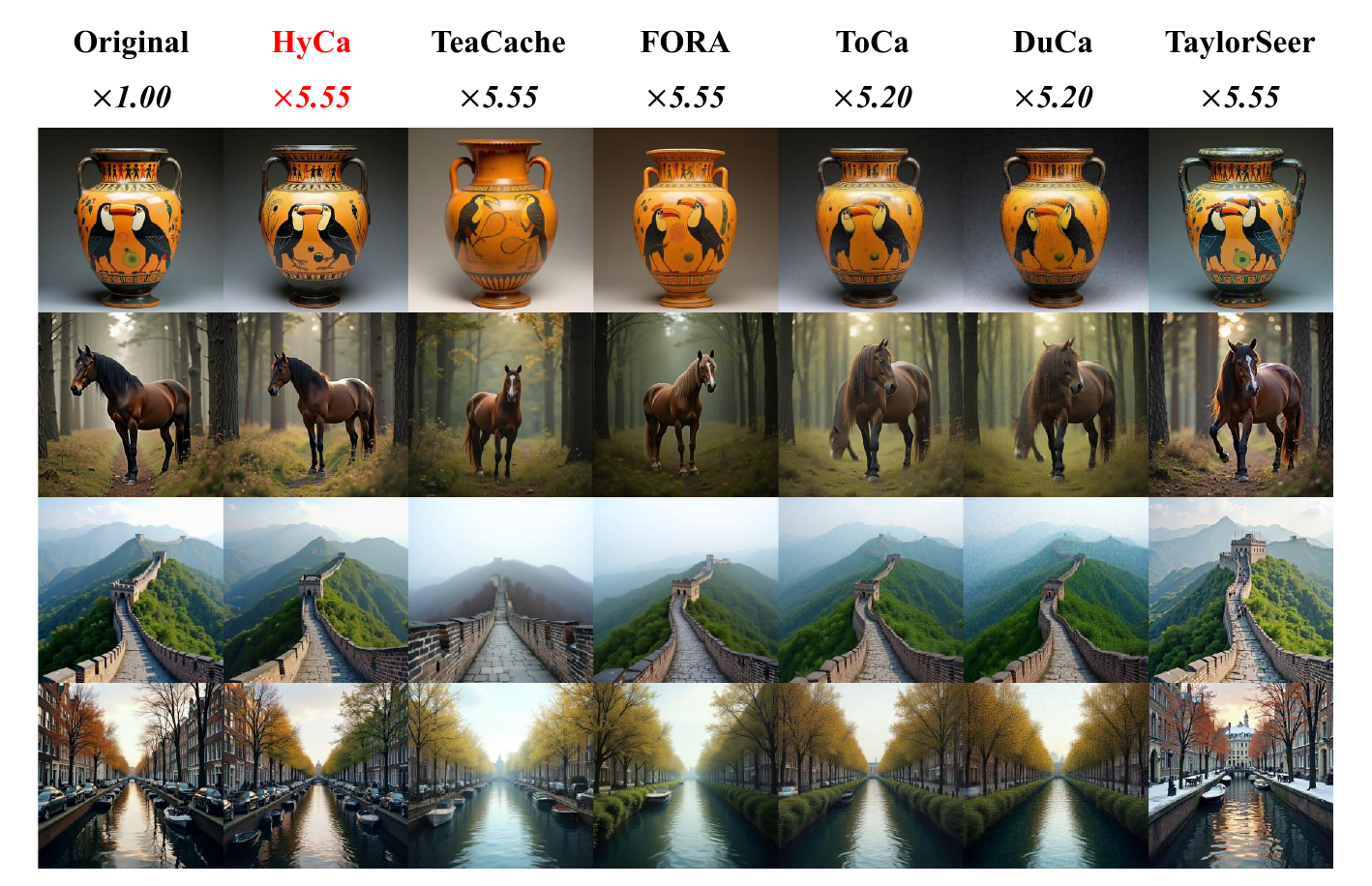}
  \caption{Visual comparison of 5.5$\times$ accelerated FLUX.}
  \vspace{-3mm}
  \label{fig:flux}
\end{figure}

On Table~\ref{table:FLUX}, HyCa consistently achieves the best speed–quality trade-off on \textbf{FLUX.1-dev}. At moderate acceleration ($\mathcal{N}{=}5$), it reaches ImageReward of \textbf{1.0066} with 4.16$\times$ FLOPs reduction, surpassing TaylorSeer (0.9857 at 3.84$\times$) and DuCa (0.9955 at 3.80$\times$). Even under aggressive settings, it maintains superior quality: \textbf{1.0014} at 5.00$\times$ and even \textbf{0.9895} at 5.55$\times$, closely matching the original model (\textbf{0.9898}) while other baselines degrade (TeaCache 0.8683, ToCa 0.7155). Visual comparison on Fig.~\ref{fig:flux} further confirms its advantage in image quality under high compression.

\noindent
\input{tab/hyvideo}

\subsection{Results on Text-to-Video Generation}

As shown in Table~\ref{table:HunyuanVideo-Metrics}, our method delivers the best performance on \textbf{Hunyuan Video}. With $\mathcal{N}{=}6$, it achieves the highest acceleration (\textbf{5.56$\times$} FLOPs reduction) while maintaining a strong VBench score (\textbf{80.25}), marginly lower than the original (\textbf{80.66}) at full 50-step inference. In contrast, TaylorSeer reaches only 4.16$\times$ with 79.93, TeaCache drops to 79.36, and DuCa/ToCa degrade further. This demonstrates a superior speed–quality trade-off and strong generalization to video generation.


\noindent
\input{tab/edit}

\subsection{Results on Image Editing}

\begin{figure}[htbp]
  \centering
  \includegraphics[trim=40 240 70 20, clip,width=1\linewidth]{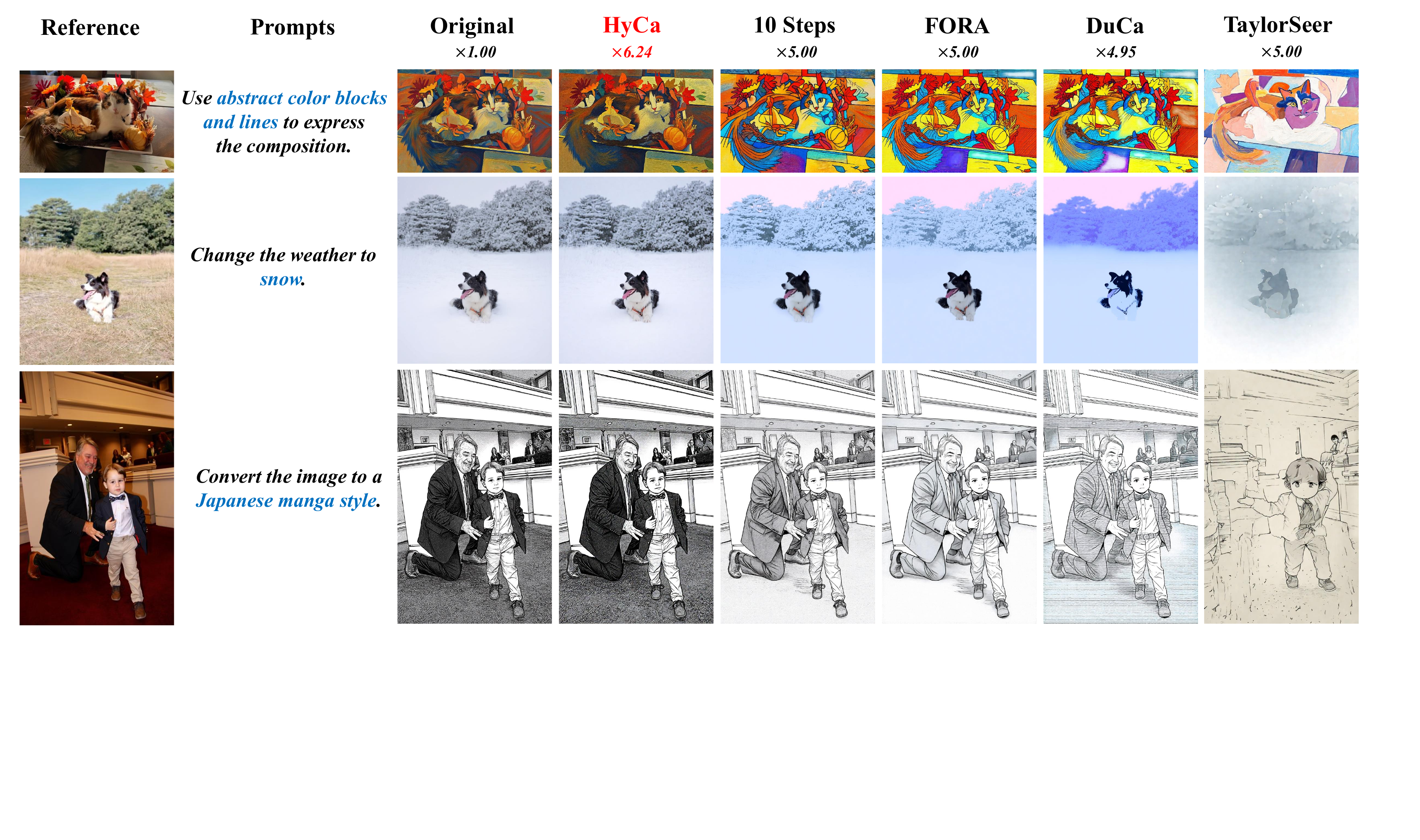}
  \caption{Visual comparison of different caching method on Qwen-Image-Edit.}
  \vspace{-3mm}
  \label{fig:edit}
\end{figure}

Our method also performs strongly on \textbf{Qwen-Image-Edit}, as shown in Table~\ref{table:qwen-edit}. At $\mathcal{N}{=}6$, it achieves the best overall scores (\textbf{7.50} CN / \textbf{7.45} EN), surpassing TaylorSeer (6.92 / 6.89), FORA (7.25 / 7.28), and even the original model's (7.41 CN). At $\mathcal{N}{=}8$, it remains leading (\textbf{7.44} CN / \textbf{7.42} EN) even exceeding the original model, while other baselines drop sharply (TaylorSeer 6.31 / 6.31). Visual comparisons in Fig.~\ref{fig:edit} further confirm HyCa’s superiority across diverse editing tasks.


\subsection{Results on Distilled Models}

To examine compatibility with distillation, we evaluate HyCa on \textbf{FLUX.1 schnell} and \textbf{Qwen-Image-Lightning}, as shown in Table~\ref{table:dis}. On FLUX.1-schnell, HyCa cuts latency from 2.34s to 1.16s for the 4-step distilled model (\textbf{24.4$\times$} over the original 50-step model) while improving quality: ImageReward increased to \textbf{0.9592} and excellent perceptual metrics (PSNR 34.37, SSIM 0.928, LPIPS 0.056). On Qwen-Image-Lightning, it reduces latency from 13.35s(8-step distilled baseline) to 6.68s (\textbf{12.2$\times$}) with quality largely preserved (ImageReward \textbf{1.2201}, CLIP 35.07) and perceptual metrics maintained (PSNR 30.97, SSIM 0.754, LPIPS 0.189). These results confirm that HyCa complements distillation, delivering extreme speedups with equal or even better quality.



\input{tab/dis}

%% file: tab/qwenimage.tex
\begin{table*}[htbp]
\centering
\vspace{-5mm}
\caption{\textbf{Quantitative comparison of text-to-image generation} on Qwen-Image.}
\vspace{-2mm}
\resizebox{\textwidth}{!}{
\begin{tabular}{l | c  c | c  c | c  c | c c c}
    \toprule
    \multirow{2}{*}{\textbf{Method}} 
    & \multicolumn{4}{c|}{\textbf{Acceleration}} 
    & \multicolumn{2}{c|}{\textbf{Quality Metrics}} 
    & \multicolumn{3}{c}{\textbf{Perceptual Metrics}} \\
    \cline{2-10}
    & \textbf{Latency(s) \(\downarrow\)} 
    & \textbf{Speed \(\uparrow\)} 
    & \textbf{FLOPs(T) \(\downarrow\)}  
    & \textbf{Speed \(\uparrow\)} 
    & \textbf{Image Reward \(\uparrow\)} 
    & \textbf{CLIP \(\uparrow\)} 
    & \textbf{PSNR \(\uparrow\)} 
    & \textbf{SSIM \(\uparrow\)} 
    & \textbf{LPIPS \(\downarrow\)}\rule{0pt}{2ex} \\
    \midrule

\textbf{Original: 50 steps }
& 74.91 & 1.00$\times$ & 12917.56 & 1.00$\times$ & 1.2547 \textcolor{gray!70}{\scriptsize (+0.000\%)} & 35.51 & $\infty$ & 1.000 & 0.000 \\

$50\%$ steps 
& 37.73 & 1.99$\times$ & 6458.78 & 2.00$\times$ & 1.2048 \textcolor{gray!70}{\scriptsize (-3.979\%)} & 35.31 & 30.85 & 0.798 & 0.249 \\

$20\%$ steps 
& 15.31 & 4.89$\times$ & 2583.51 & 5.00$\times$ & 0.9234 \textcolor{gray!70}{\scriptsize (-26.42\%)} & 35.17 & 28.52 & 0.627 & 0.504 \\
\midrule

\textbf{TaylorSeer}($\mathcal{N}=3$)
& 36.90 & 2.03$\times$ & 4646.60 & 2.78$\times$ & 1.0685 \textcolor{gray!70}{\scriptsize (-14.83\%)} & 34.76 & 28.29 & 0.504 & 0.628 \\

\rowcolor{gray!20}
\textbf{HyCa}($\mathcal{N}=3$)
& \textbf{35.33} & \textbf{2.12}$\times$ & \textbf{4646.60} & \textbf{2.78}$\times$ & \bf {1.2363} \textcolor[HTML]{0f98b0}{\scriptsize (-1.465\%)} & \textbf{34.97} & \textbf{30.42} & \textbf{0.763} & \textbf{0.247} \\

\midrule

\textbf{FORA}($\mathcal{N}=5$)
& 21.71 & 3.45$\times$ & 2585.46 & 5.00$\times$ & 0.7767 \textcolor{gray!70}{\scriptsize (-38.10\%)} & 34.47 & 24.55 & 0.553 & 0.556 \\

\textbf{\texttt{ToCa}}($\mathcal{N}=8$)
& 60.62 & 1.24$\times$ & 2991.34 & 4.32$\times$ & 0.9673 \textcolor{gray!70}{\scriptsize (-22.87\%)} & 34.83 & 29.00 & 0.643 & 0.417 \\

\textbf{\texttt{DuCa}}($\mathcal{N}$=9)
& 24.83 & 3.02$\times$ & 2958.13 & 4.37$\times$ & 0.8213 \textcolor{gray!70}{\scriptsize (-34.53\%)} & {34.69} &{28.42} & {0.582} & {0.531} \\

\textbf{TaylorSeer}($\mathcal{N}=6$)
& 24.61 & 3.04$\times$ & 2585.46 & 5.00$\times$ & 0.9483 \textcolor{gray!70}{\scriptsize (-24.41\%)} & 34.76 & 28.29 & 0.504 & 0.628 \\

\rowcolor{gray!20}
\textbf{HyCa}($\mathcal{N}=6$)
& \textbf{21.58} & \textbf{3.47}$\times$ & \textbf{2584.46} & \textbf{5.00}$\times$ & \bf {1.1939} \textcolor[HTML]{0f98b0}{\scriptsize (-4.848\%)} & \textbf{34.87} & \textbf{29.65} & \textbf{0.709} & \textbf{0.320} \\

\midrule

\textbf{FORA}($\mathcal{N}=6$)
& 17.89 & 4.19$\times$ & 2323.30 & 5.56$\times$ & 0.4781 \textcolor{gray!70}{\scriptsize (-61.91\%)} & 28.50 & 28.38 & 0.546 & 0.597 \\

\textbf{\texttt{ToCa}}($\mathcal{N}=12$)
& 52.72 & 1.42$\times$ & 2406.20 & 5.37$\times$ & 0.5593 \textcolor{gray!70}{\scriptsize (-55.42\%)} & 33.92 & 28.72 & 0.589 & 0.519 \\

\textbf{\texttt{DuCa}}($\mathcal{N}$=12)
& 21.82 & 3.43$\times$ & 2171.56 & 5.95$\times$ & 0.5225 \textcolor{gray!70}{\scriptsize (-58.34\%)} & 33.97 & 28.37 & 0.576 & 0.593 \\

\textbf{TaylorSeer}($\mathcal{N}=7$)
& 21.88 & 4.32$\times$ & 2323.30 & 5.56$\times$ & 0.9113 \textcolor{gray!70}{\scriptsize (-27.39\%)} & 34.30 & 28.20 & 0.481 & 0.652 \\

\rowcolor{gray!20}
\textbf{HyCa}($\mathcal{N}=8$)
& \textbf{13.92} & \textbf{5.38}$\times$ & \textbf{2066.81} & \textbf{6.25}$\times$ & \bf {1.0811} \textcolor[HTML]{0f98b0}{\scriptsize (-13.84\%)} & \textbf{34.75} & \textbf{28.89} & \textbf{0.600} & \textbf{0.433} \\

\bottomrule
\end{tabular}}
\label{table:Qwenimage}
\end{table*}

%% file: tab/flux_full.tex
\begin{table*}[htbp]
\centering
\vspace{-7mm}
\caption{\textbf{Quantitative comparison of text-to-image generation} on FLUX.1-dev.}
\vspace{-2mm}
\setlength\tabcolsep{7pt} 
\small
\resizebox{\textwidth}{!}{
\begin{tabular}{l | c | c  c | c  c | c | c }
    \toprule
    \rowcolor{white}
  {\bf Method} & {\bf Efficient} & \multicolumn{4}{c|}{\bf Acceleration} & \multirow{2}{*} {\bf Image Reward $\uparrow$} & \multirow{2}{*}{\bf CLIP Score $\uparrow$} \\
    \cline{3-6}
    {\bf FLUX.1} & {\bf Attention } & {\bf Latency(s) $\downarrow$} & {\bf Speed $\uparrow$} & {\bf FLOPs(T) $\downarrow$}  & {\bf Speed $\uparrow$}\rule{0pt}{2ex} &  & \\
    \midrule
        $\textbf{[dev]: 50 steps}$ & \ding{52} & 25.82 & 1.00$\times$ & 3719.50 & 1.00$\times$ & 0.9898 \textcolor{gray!70}{\scriptsize (+0.000\%)} & 32.404 \textcolor{gray!70}{\scriptsize (+0.000\%)} \\ 
        \midrule

        {$60\%$\textbf{ steps}} & \ding{52} & 16.70 & 1.55$\times$ & 2231.70 & 1.67$\times$ & 0.9663 \textcolor{gray!70}{\scriptsize (-2.371\%)} & 32.312 \textcolor{gray!70}{\scriptsize (-0.283\%)} \\
        {$\Delta$-DiT} ($\mathcal{N}=2$) & \ding{52} & 17.80 & 1.45$\times$ & 2480.01 & 1.50$\times$ & 0.9444 \textcolor{gray!70}{\scriptsize (-4.594\%)} & 32.273 \textcolor{gray!70}{\scriptsize (-0.404\%)} \\
        {$\Delta$-DiT} ($\mathcal{N}=3$) & \ding{52} & 13.02 & 1.98$\times$ & 1686.76 & 2.21$\times$ & 0.8721 \textcolor{gray!70}{\scriptsize (-11.90\%)} & 32.102 \textcolor{gray!70}{\scriptsize (-0.933\%)} \\
        $\textbf{DBcache}$ & \ding{52} & 16.88 & 1.53$\times$ & 2384.29 & 1.56$\times$ & 1.0069 \textcolor{gray!70}{\scriptsize (+1.725\%)} & 32.530 \textcolor{gray!70}{\scriptsize (+0.389\%)} \\
        $\textbf{TaylorSeer}$ $(\mathcal{N}=3, O=2)$ & \ding{52} & 9.89 & 2.61$\times$ & 1320.07 & 2.82$\times$ & 0.9989 \textcolor{gray!70}{\scriptsize (+0.919\%)} & 32.413 \textcolor{gray!70}{\scriptsize (+0.027\%)} \\
        $\textbf{FoCa}$ $(\mathcal{N}=3)$ & \ding{52} & 9.28 & 2.78$\times$ & 1327.21 & 2.80$\times$ & 0.9890 \textcolor{gray!70}{\scriptsize (-0.081\%)} & 32.577 \textcolor{gray!70}{\scriptsize (+0.533\%)} \\
        \rowcolor{gray!20}
        $\textbf{HyCa}$ $(\mathcal{N}=4)$ & \ding{52} & \bf 8.09 & \textbf{3.19$\times$} & \bf 967.91 & \textbf{3.84$\times$} &\bf 1.0182 \textcolor[HTML]{0f98b0}{\scriptsize (+2.865\%)} & \bf 32.671 \textcolor[HTML]{0f98b0}{\scriptsize (+0.822\%)} \\
        \midrule

        {$34\%$\textbf{ steps}} & \ding{52} & 9.07 & 2.85$\times$ & 1264.63 & 3.13$\times$ & 0.9453 \textcolor{gray!70}{\scriptsize (-4.498\%)} & 32.114 \textcolor{gray!70}{\scriptsize (-0.893\%)} \\
        $\textbf{Chipmunk}$ & \ding{52} & 12.72 & 2.02$\times$ & 1505.87 & 2.47$\times$ & 0.9936 \textcolor{gray!70}{\scriptsize (+0.384\%)} & 32.548 \textcolor{gray!70}{\scriptsize (+0.444\%)} \\
        $\textbf{FORA}$ $(\mathcal{N}=3)$ & \ding{52} & 10.16 & 2.54$\times$ & 1320.07 & 2.82$\times$ & 0.9776 \textcolor{gray!70}{\scriptsize (-1.232\%)} & 32.266 \textcolor{gray!70}{\scriptsize (-0.425\%)} \\
        $\textbf{\texttt{ToCa}}$ $(\mathcal{N}=6)$ & \ding{56} & 13.16 & 1.96$\times$ & 924.30 & 4.02$\times$ & 0.9802 \textcolor{gray!70}{\scriptsize (-0.968\%)} & 32.083 \textcolor{gray!70}{\scriptsize (-0.990\%)} \\
        $\textbf{\texttt{DuCa}}$ $(\mathcal{N}=5)$ & \ding{52} & 8.18 & 3.15$\times$ & 978.76 & 3.80$\times$ & 0.9955 \textcolor{gray!70}{\scriptsize (+0.576\%)} & 32.241 \textcolor{gray!70}{\scriptsize (-0.503\%)} \\
        $\textbf{TaylorSeer}$ $(\mathcal{N}=4, O=2)$ & \ding{52} & 9.24 & 2.80$\times$ & 967.91 & 3.84$\times$ & 0.9857 \textcolor{gray!70}{\scriptsize (-0.414\%)} & 32.413 \textcolor{gray!70}{\scriptsize (+0.027\%)} \\
        $\textbf{FoCa}$ $(\mathcal{N}=4)$ & \ding{52} & 9.35 & 2.76$\times$ & 1050.70 & 3.54$\times$ & 0.9757 \textcolor{gray!70}{\scriptsize (-1.424\%)} & 32.538 \textcolor{gray!70}{\scriptsize (+0.414\%)} \\
        $\textbf{Clusca}$ $(\mathcal{N}=4, O=2, K=16)$ & \ding{52} & 9.25 & 2.79$\times$ & 1045.58 & 3.56$\times$ & 0.9850 \textcolor{gray!70}{\scriptsize (-0.485\%)} & 32.441 \textcolor{gray!70}{\scriptsize (+0.114\%)} \\
        \rowcolor{gray!20}
        $\textbf{HyCa}$ $(\mathcal{N}=5)$ & \ding{52} & \bf 7.62 & \textbf{3.38$\times$} & \bf 893.54 & \textbf{4.16$\times$} &\bf 1.0066 \textcolor[HTML]{0f98b0}{\scriptsize (+1.700\%)} &\bf 32.693 \textcolor[HTML]{0f98b0}{\scriptsize (+0.890\%)} \\
        \midrule

        $\textbf{FORA}$ $(\mathcal{N}=4)$ & \ding{52} & 8.12 & 3.14$\times$ & 967.91 & 3.84$\times$ & 0.9730 \textcolor{gray!70}{\scriptsize (-1.695\%)} & 32.142 \textcolor{gray!70}{\scriptsize (-0.809\%)} \\
        $\textbf{\texttt{ToCa}}$ $(\mathcal{N}=8)$ & \ding{56} & 11.36 & 2.27$\times$ & 784.54 & 4.74$\times$ & 0.9451 \textcolor{gray!70}{\scriptsize (-4.514\%)} & 31.993 \textcolor{gray!70}{\scriptsize (-1.271\%)} \\
        $\textbf{\texttt{DuCa}}$ $(\mathcal{N}=7)$ & \ding{52} & \bf6.74 & \textbf{3.83$\times$} & 760.14 & 4.89$\times$ & 0.9757 \textcolor{gray!70}{\scriptsize (-1.424\%)} & 32.066 \textcolor{gray!70}{\scriptsize (-1.046\%)} \\
        \textbf{TeaCache} $({l}=0.8)$ & \ding{52} & 7.21 & 3.58$\times$ & 892.35 & 4.17$\times$ & 0.8683 \textcolor{gray!70}{\scriptsize (-12.28\%)} & 31.704 \textcolor{gray!70}{\scriptsize (-2.159\%)} \\
        $\textbf{TaylorSeer}$ $(\mathcal{N}=5, O=2)$ & \ding{52} & 7.46 & 3.46$\times$ & 893.54 & 4.16$\times$ & 0.9768 \textcolor{gray!70}{\scriptsize (-1.314\%)} & 32.467 \textcolor{gray!70}{\scriptsize (+0.194\%)} \\
        $\textbf{FoCa}$ $(\mathcal{N}=6)$ & \ding{52} & 7.54 & 3.42$\times$ & 745.39 & 4.99$\times$ & 0.9713 \textcolor{gray!70}{\scriptsize (-1.870\%)} & \textbf{32.922} \textcolor{gray!70}{\scriptsize (+1.600\%)} \\
        $\textbf{Speca}$ $(\mathcal{N}_{\text{max}}=8, \mathcal{N}_{\text{min}}=2)$ & \ding{52} & 7.42 & 3.48$\times$ & 791.38 & 4.70$\times$ & 0.9985 \textcolor{gray!70}{\scriptsize (+0.878\%)} & 32.277 \textcolor{gray!70}{\scriptsize (-0.391\%)} \\
        $\textbf{Clusca}$ $(\mathcal{N}=5, O=1, K=16)$ & \ding{52} & 7.05 & 3.66$\times$ & 897.03 & 4.14$\times$ & 0.9718 \textcolor{gray!70}{\scriptsize (-1.818\%)} & 32.319 \textcolor{gray!70}{\scriptsize (-0.262\%)} \\
        \rowcolor{gray!20}
        $\textbf{HyCa}$ $(\mathcal{N}=6)$ & \ding{52} &  6.81 & {3.79$\times$} & \bf 744.81 & \textbf{5.00$\times$} &\bf 1.0014 \textcolor[HTML]{0f98b0}{\scriptsize (+1.163\%)} & 32.483 \bf \textcolor[HTML]{0f98b0}{\scriptsize (+0.244\%)} \\
        \midrule

        $\textbf{FORA}$ $(\mathcal{N}=6)$ & \ding{52} & 8.17 & 3.16$\times$ & 744.80 & 4.99$\times$ & 0.7760 \textcolor{gray!70}{\scriptsize (-21.62\%)} & 31.742 \textcolor{gray!70}{\scriptsize (-2.043\%)} \\
        $\textbf{\texttt{ToCa}}$ $(\mathcal{N}=10)$ & \ding{56} & 7.93 & 3.25$\times$ & 714.66 & 5.20$\times$ & 0.7155 \textcolor{gray!70}{\scriptsize (-27.70\%)} & 31.808 \textcolor{gray!70}{\scriptsize (-1.839\%)} \\
        $\textbf{\texttt{DuCa}}$ $(\mathcal{N}=9)$ & \ding{52} & 7.27 & 3.55$\times$ & 690.25 & 5.39$\times$ & 0.8382 \textcolor{gray!70}{\scriptsize (-15.33\%)} & 31.759 \textcolor{gray!70}{\scriptsize (-1.993\%)} \\
        \textbf{TeaCache} $({l}=1)$ & \ding{52} & 8.19 & 3.19$\times$ & 743.63 & 5.01$\times$ & 0.8379 \textcolor{gray!70}{\scriptsize (-15.36\%)} & 31.877 \textcolor{gray!70}{\scriptsize (-1.627\%)} \\
        $\textbf{TaylorSeer}$ $(\mathcal{N}=7, O=2)$ & \ding{52} & 6.77 & 3.81$\times$ & 671.39 & 5.54$\times$ & 0.9698 \textcolor{gray!70}{\scriptsize (-2.020\%)} & 32.128 \textcolor{gray!70}{\scriptsize (-0.851\%)} \\
        $\textbf{Clusca}$ $(\mathcal{N}=6, O=1, K=16)$ & \ding{52} & 7.13 & 3.62$\times$ & 748.48 & 4.97$\times$ & 0.9704 \textcolor{gray!70}{\scriptsize (-1.956\%)} & 32.217 \textcolor{gray!70}{\scriptsize (-0.577\%)} \\
        \rowcolor{gray!20}
        $\textbf{HyCa}$ $(\mathcal{N}=7)$ & \ding{52} & \bf 6.43 & \textbf{4.01$\times$} & \bf 670.44 & \textbf{5.55$\times$} &\bf 0.9895 \textcolor[HTML]{0f98b0}{\scriptsize (-0.030\%)} &\bf 32.520 \textcolor[HTML]{0f98b0}{\scriptsize (+0.358\%)} \\
\bottomrule
\end{tabular}}
\label{table:FLUX}
\footnotesize
\end{table*}

%% file: tab/hyvideo.tex
\begin{table*}[htb]
\centering

\vspace{-4mm}
\caption{\textbf{Quantitative comparison of text-to-video generation} on HunyuanVideo.}
\vspace{-2mm}
\setlength\tabcolsep{5.0pt} 
\small
\resizebox{\textwidth}{!}{
\begin{tabular}{l | c | c  c | c  c | c }
    \toprule
    {\bf Method} & {\bf Efficient} &\multicolumn{4}{c|}{\bf Acceleration} &{\bf VBench $\uparrow$}  \\
    \cline{3-6}
    {\bf HunyuanVideo} & {\bf Attention } & {\bf Latency(s) $\downarrow$} & {\bf Speed $\uparrow$} & {\bf FLOPs(T) $\downarrow$}  & {\bf Speed $\uparrow$} & \bf Score(\%)\rule{0pt}{2ex}\\ 
    \midrule

  $\textbf{Original: 50 steps}$ 
                           & \ding{52}  & 145.00 & 1.00$\times$& {29773.0}   & {1.00$\times$} & 80.66 \textcolor{gray!70}{\scriptsize (+0.0\%)}      \\ 
    \midrule
  
  {$22\%$\textbf{ steps}}  & \ding{52}  & 31.87 & 4.55$\times$ & {6550.1}   & {4.55$\times$} & 78.74 \textcolor{gray!70}{\scriptsize (-2.4\%)}          \\

\textbf{TeaCache}$({l}=0.4)$ 
                           & \ding{52} & 30.49 & 4.76$\times$ & 6550.1 & 4.55$\times$ & 79.36 \textcolor{gray!70}{\scriptsize (-1.6\%)}  \\

$\textbf{FORA}(N=5)$ 
& \ding{52}  & 34.39 & 4.22$\times$  & {5960.4}   & {5.00$\times$} & 78.83 \textcolor{gray!70}{\scriptsize (-2.3\%)}     \\

$\textbf{\texttt{ToCa}}$ $(\mathcal{N}=5,R=90\%)$
                           & \ding{56}  & 38.52 & 3.76$\times$ & {7006.2}   & {4.25$\times$} & 78.86 \textcolor{gray!70}{\scriptsize (-2.2\%)}    \\
  
$\textbf{\texttt{DuCa}} $ $(\mathcal{N}=5,R=90\%)$ 
                           & \ding{52}  & 31.69 & 4.58$\times$ & {6483.2}   & {4.48$\times$} & 78.72 \textcolor{gray!70}{\scriptsize (-2.4\%)}    \\
                           
$\textbf{TaylorSeer} $ $(\mathcal{N}=5,O=1)$ 
                           & \ding{52}  & 34.84 & 4.16$\times$ & {5960.4}  & {5.00$\times$} & 79.93 \textcolor{gray!70}{\scriptsize (-0.9\%)} \\

$\textbf{Speca} $ $(\mathcal{N}_{\text{max}}=8, \mathcal{N}_{\text{min}}=2)$ 
                           & \ding{52}  & 34.58 & 4.19$\times$ & {5692.7}  & {5.23$\times$} & 79.98 \textcolor{gray!70}{\scriptsize (-0.8\%)} \\

$\textbf{Clusca} $ $(\mathcal{N}=5,O=1,K=16)$ 
                           & \ding{52}  & 35.37 & 4.10$\times$ & {5373.0}  & {5.54$\times$} & 79.96 \textcolor{gray!70}{\scriptsize (-0.9\%)} \\

$\textbf{FoCa} $ $(\mathcal{N}=5)$ 
                           & \ding{52}  & 34.52 & 4.20$\times$ & {5966.5}  & {4.99$\times$} & 79.96 \textcolor{gray!70}{\scriptsize (-0.8\%)} \\

\rowcolor{gray!20}
$\textbf{HyCa} $ $(\mathcal{N}=6)$ 
                           & \ding{52}  & \bf 28.48 & \textbf{5.09$\times$} & \bf 5359.1  & \textbf{5.56$\times$} & \bf {80.25} \textcolor[HTML]{0f98b0}{\scriptsize (-0.5\%)}  \\

    \bottomrule
\end{tabular}}
\label{table:HunyuanVideo-Metrics}
\end{table*}

%% file: tab/edit.tex
\begin{table*}[htbp]
\centering
\vspace{-4mm}
\caption{\textbf{Quantitative comparison of image editing} on Qwen-Image-Edit.}
\vspace{-2mm}
\resizebox{\textwidth}{!}{ 
\begin{tabular}{l | c  c | c  c | c c c|c c c}
\toprule
\multirow[c]{2}{*}{\bf Method}
& \multicolumn{4}{c|}{\bf Acceleration} 
& \multicolumn{3}{c|}{\bf GEdit-CN (Full)}
& \multicolumn{3}{c}{\bf GEdit-EN (Full)} \rule{0pt}{2ex}\\
\cline{2-11}
& {\bf Latency(s) $\downarrow$} 
& {\bf Speed $\uparrow$} 
& {\bf FLOPs(T) $\downarrow$}  
& {\bf Speed $\uparrow$} 
& {\bf SC $\uparrow$} 
& {\bf PQ $\uparrow$} 
& {\bf OS $\uparrow$}
& {\bf SC $\uparrow$} 
& {\bf PQ $\uparrow$} 
& {\bf OS $\uparrow$}\rule{0pt}{2ex}\\
\midrule

\textbf{Original: 50 steps}
& 284.51 & 1.00$\times$ & 28190.88 & 1.00$\times$ &  7.68 &  7.51 &  7.41 &  7.82 &  7.54 &  7.54 \\

$50\%$ steps
& 143.29 & 1.99$\times$ & 14095.44 & 2.00$\times$ & 7.70 & 7.53 & 7.44 & 7.77 & 7.52 & 7.47 \\

$20\%$ steps
& 58.45 & 4.87$\times$ & 5638.18 & 5.00$\times$ & 7.65 & 7.42 & 7.35 & 7.73 & 7.46 & 7.44 \\

\midrule
\textbf{FORA}($\mathcal{N}=5$)
& 63.15 & 4.51$\times$ & 5643.13 & 5.00$\times$ & 7.60 & 7.31 & 7.25 & 7.62 & 7.34 & 7.28 \\

$\textbf{\texttt{DuCa}} $ $(\mathcal{N}=6,R=90\%)$ & 70.95 & 4.01$\times$ & 5897.67 & 4.78$\times$ & 7.63 & 7.44 & 7.44 & 7.68 & 7.42 & 7.39 \\

\textbf{TaylorSeer}($\mathcal{N}=6$)
& 65.66 & 4.33$\times$ & 5643.13 & 5.00$\times$ & 7.25 & 7.09 & 6.92 & 7.26 & 7.14 & 6.89 \\

\rowcolor{gray!20}
\textbf{HyCa} ($\mathcal{N}=6$)
& \bf 62.89 & \textbf{4.52}$\times$ & \bf 5642.24 & \textbf{5.00$\times$} & \textbf{7.76} & \textbf{7.49} & \textbf{7.50} & \textbf{7.77} & \textbf{7.47} & \textbf{7.45} \\

\midrule

\textbf{FORA}($\mathcal{N}=7$)
& 52.20 & 5.45$\times$ & 4515.74 & 6.24$\times$ & 7.42 & 7.13 & 7.06 & 7.43 & 7.19 & 7.06 \\

$\textbf{\texttt{DuCa}} $ $(\mathcal{N}=10,R=95\%)$ & 59.81 & 4.76$\times$ & 5158.45 & 5.46$\times$ & 7.50 & 5.75 & 6.39 & 7.52 & 5.77 & 6.41 \\

\textbf{TaylorSeer}($\mathcal{N}=8$)
& 53.92 & 5.28$\times$ & 4515.74 & 6.24$\times$ & 6.61 & 6.65 & 6.31 & 6.67 & 6.63 & 6.31 \\

\rowcolor{gray!20}
\textbf{HyCa} ($\mathcal{N}=8$)
& \bf 51.09 & \textbf{5.57}$\times$ & \bf 4514.48 & \textbf{6.24$\times$} & \textbf{7.74} & \textbf{7.41} & \textbf{7.44} & \textbf{7.80} & \textbf{7.36} & \textbf{7.42} \\

\bottomrule
\end{tabular}
}
\label{table:qwen-edit}
\raggedright

{\scriptsize
\begin{itemize}[leftmargin=10pt,topsep=0pt]
\item SC denotes Semantic Consistency on Gedit Bench, PQ denotes Perceptual Quality, OS denotes the Overall Score.
\end{itemize}
}

\vspace{-6mm}
\end{table*}

%% file: tab/dis.tex
\begin{table*}[htbp]
\centering
\caption{\textbf{Quantitative comparison of Distilled Model} on Flux and Qwen-Image.}
\vspace{-2mm}
\resizebox{\textwidth}{!}{
\begin{tabular}{l | c  c | c  c | c  c | c c c}
    \toprule
    \multirow{2}{*}{\textbf{Method}} 
    & \multicolumn{4}{c|}{\textbf{Acceleration}} 
    & \multicolumn{2}{c|}{\textbf{Quality Metrics}} 
    & \multicolumn{3}{c}{\textbf{Perceptual Metrics}} \\
    \cline{2-10}
    & \textbf{Latency(s) \(\downarrow\)} 
    & \textbf{Speed \(\uparrow\)} 
    & \textbf{FLOPs(T) \(\downarrow\)}  
    & \textbf{Speed \(\uparrow\)} 
    & \textbf{ImageReward\(\uparrow\)} 
    & \textbf{CLIP\(\uparrow\)} 
    & \textbf{PSNR\(\uparrow\)} 
    & \textbf{SSIM\(\uparrow\)} 
    & \textbf{LPIPS\(\downarrow\)}\rule{0pt}{2ex} \\
    \midrule

\textbf{FLUX.1[dev]: 50 steps} 
& 25.82 & 1.00$\times$ & 3719.50 & 1.00$\times$ & 0.9898 \textcolor{gray!70}{\scriptsize (+8.380\%)} & 32.40 & - & - & - \\

\textbf{FLUX.1[schnell]: 4 steps} 
& 2.34 & 11.03$\times$ & 297.60 & 12.50$\times$ & 0.9133 \textcolor{gray!70}{\scriptsize (+0.000\%)} & 33.85 & $\infty$ & 1.000 & 0.000 \\

\textbf{TaylorSeer} ($\mathcal{N}=2$): 4 steps
& 1.58 & 16.34$\times$ & 209.70 & 17.74$\times$ & 0.9191 \textcolor{gray!70}{\scriptsize (+0.636\%)} & 33.76 & 29.13 & 0.746 & 0.249 \\

\textbf{TeaCache} $({l}=0.6)$: 4 steps
& 1.26 & 20.49$\times$ & 163.78 & 22.71$\times$ & 0.9023 \textcolor{gray!70}{\scriptsize (-1.210\%)} & 33.87 & 28.01 & 0.379 & 0.734 \\

\rowcolor{gray!20}
\textbf{HyCa ($\mathcal{N}=2$): 4 steps} 
& \textbf{1.16} & \textbf{22.25$\times$} & \textbf{152.32} & \textbf{24.42$\times$} & \bf {0.9592} \textcolor[HTML]{0f98b0}{\scriptsize (+5.029\%)} & \textbf{34.18} & \textbf{34.37} & \textbf{0.928} & \textbf{0.056} \\

\midrule

\textbf{Qwen-Image: 50 steps} 
& 74.91 & 1.00$\times$ & 12917.56 & 1.00$\times$ & 1.2547 \textcolor{gray!70}{\scriptsize (+0.232\%)} & 35.51 & - & - & - \\

\textbf{Qwen-Image-Lightning: 8 steps} 
& 13.35 & 5.61$\times$ & 2113.67 & 6.11$\times$ & 1.2518 \textcolor{gray!70}{\scriptsize (+0.000\%)} & 35.32 & $\infty$ & 1.000 & 0.000 \\

\textbf{TaylorSeer} ($\mathcal{N}=2$): 8 steps
& 8.11 & 9.24$\times$ & 1320.81 & 9.78$\times$ & 1.0418 \textcolor{gray!70}{\scriptsize (-16.79\%)} & 34.44 & 29.49 & 0.62 & 0.377 \\

\textbf{TaylorSeer} ($\mathcal{N}=3$): 8 steps
& 6.78 & 11.07$\times$ & 1057.08 & 12.22$\times$ & 0.2644 \textcolor{gray!70}{\scriptsize (-78.89\%)} & 30.55 & 27.98 & 0.30 & 0.672 \\

\rowcolor{gray!20}
\textbf{HyCa ($\mathcal{N}=2$): 8 steps} 
& 8.20 & 9.13$\times$ & 1320.81 & 9.78$\times$ & \bf  {1.2478} \textcolor[HTML]{0f98b0}{\scriptsize (-0.320\%)} & \textbf{35.27} & \textbf{32.52} & \textbf{0.837} & \textbf{0.119} \\

\rowcolor{gray!20}
\textbf{HyCa ($\mathcal{N}=3$): 8 steps} 
& \textbf{6.68} & \textbf{11.21$\times$} & \textbf{1057.08} & \textbf{12.22$\times$} & \bf 1.2201 \textcolor[HTML]{0f98b0}{\scriptsize (-2.542\%)} & 35.07 & 30.97 & 0.754 & 0.189 \\

\bottomrule
\end{tabular}}

{\scriptsize
\begin{itemize}[leftmargin=10pt,topsep=0pt]
\item The PSNR, SSIM, and LPIPS of HyCa are computed with respect to the outputs of the corresponding distilled baseline models.
\end{itemize}
}

\vspace{-3mm}
\label{table:dis}
\end{table*}

%% file: sec/6_discussion.tex
\begin{figure}[htbp]
  \centering
  \includegraphics[trim=150 20 150 40, clip,width=1\linewidth]{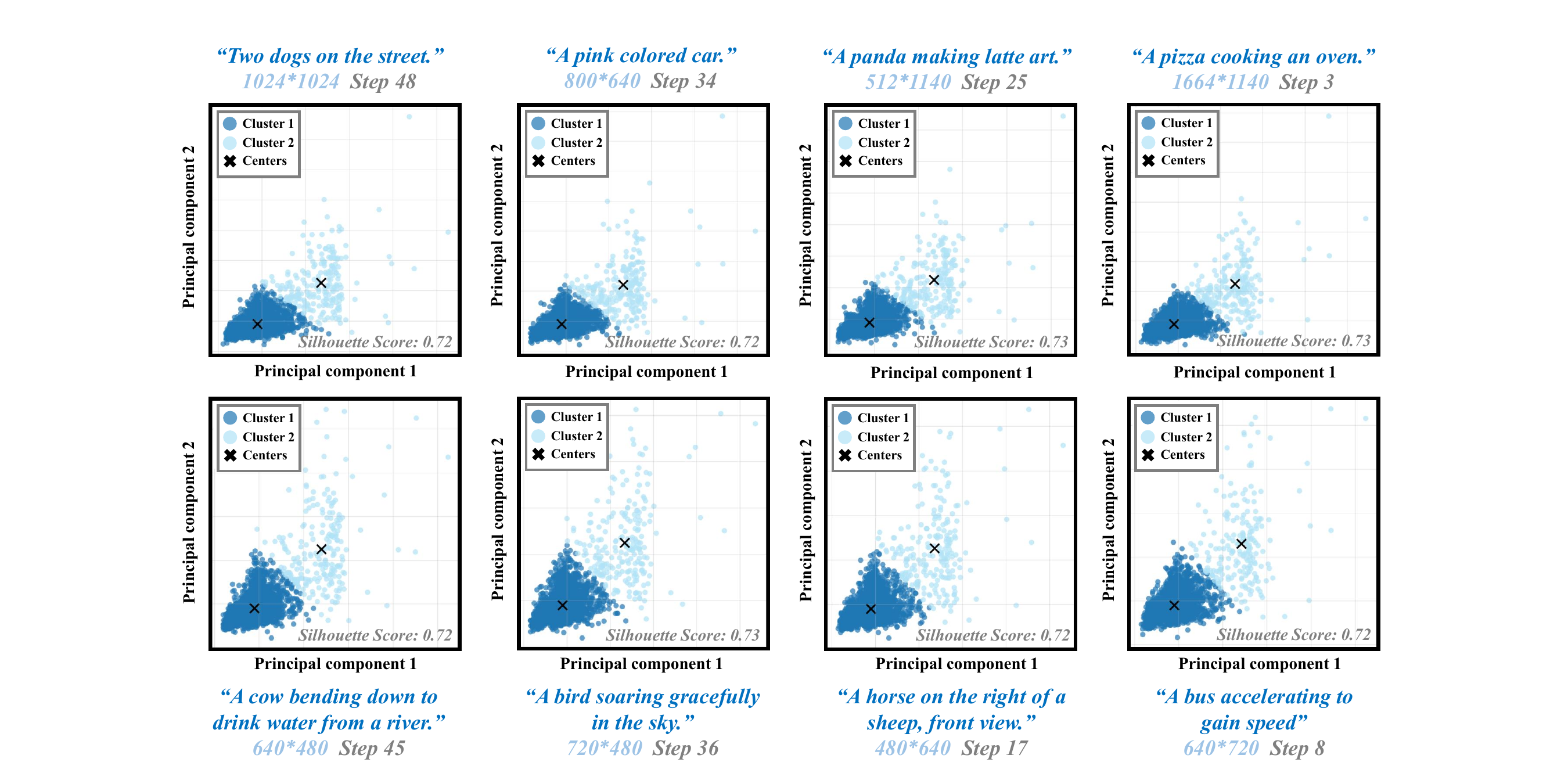}
  \caption{Top row: Clustering results from FLUX.1dev; Bottom row: Clustering results from Hunyuan Video. The clustering assignments remain highly consistent across various prompts, resolutions and timesteps, suggesting stable and robust geometric structure in the feature space.}
   \vspace{-5mm}
  \label{fig:cluster}
  
\end{figure}

\section{Discussion}

\begin{figure}[htbp]
  \centering
  \includegraphics[trim=105 90 130 135, clip,width=1\linewidth]{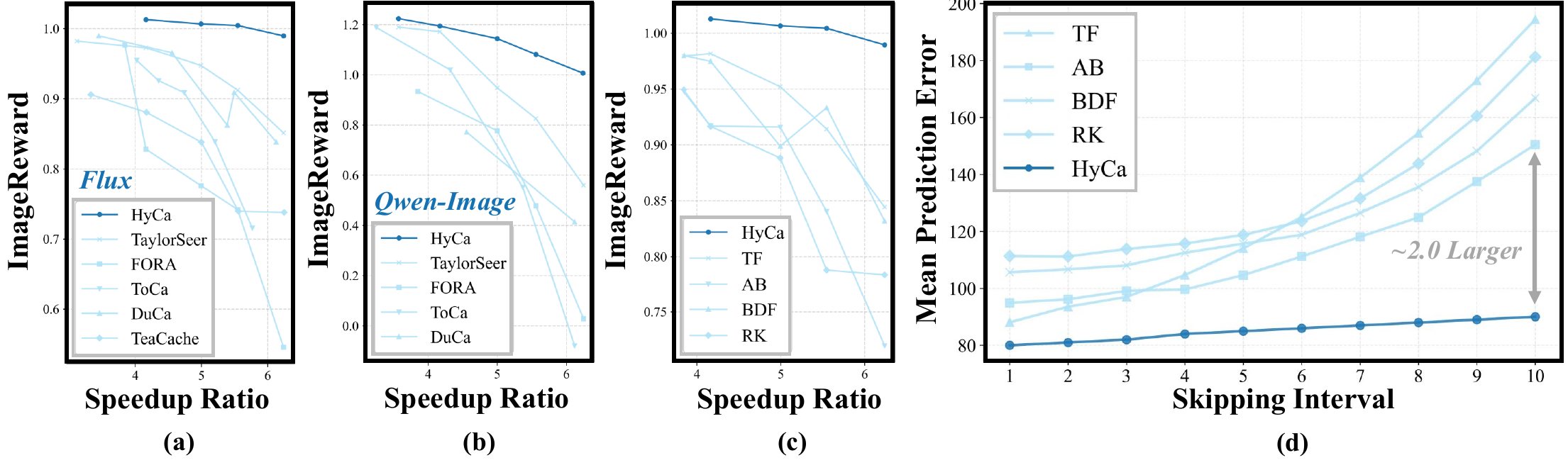}
   \vspace{-3mm}
  \caption{
\textbf{Overall and ablation results of HyCa.} 
(a--b) HyCa consistently outperforms token-wise (ToCa) and one-size-fits-all (Taylorseer) baselines. 
(c--d) Ablation on FLUX shows HyCa surpasses all single-solver baselines in the solver pool, maintaining lower error and better quality.
}
 \vspace{-6mm}
  \label{fig:ablation}
\end{figure}

\noindent\textbf{Ablation Study.}
We conduct our ablation study on FLUX, as shown in Fig.~\ref{fig:ablation} (c)(d), HyCa consistently achieves higher ImageReward and lower prediction error than any individual solver baselines from our solver pool under the same conditions. These results confirm that HyCa benefits from combining diverse solvers rather than relying on a single integration strategy.

\noindent\textbf{Why Dimension-Wise Assignment?} A central design in HyCa is to assign caching strategies at feature-dimension level rather than token level primarily due to its better stability, as shown in Fig~\ref{fig:cluster}, clustering results in feature space remain nearly invariant once identified. Thus, solver assignments can be reused extensively, reducing both computational and data requirements. On the contrary, token-wise assignment varies with prompt and resolution, requiring frequent re-selection during inference. Empirically, Fig.~\ref{fig:ablation} (a)(b) confirms that our dimension-wise assignment outperforms both token-wise (ToCa, DuCa) and one-size-fits-all (FORA, TaylorSeer) baselines.

\noindent\textbf{Compatibility with Distillation.} 
Feature caching is traditionally difficult to adapt to distilled models, as distillation drastically reduces sampling steps (e.g., from 50 to 4 or 8), making feature trajectories more discrete and oscillatory. Prior caching methods rely heavily on smooth temporal evolution and thus fail in this setting. In contrast, HyCa remains effective: its solver pool includes implicit methods suited for discrete or oscillatory dynamics, and solvers are assigned per cluster for each model. This flexibility makes HyCa fully compatible with distillation, achieving substantial acceleration while preserving generation quality.

%% file: sec/7_conclusion.tex
\section{Conclusion}


We introduced \textbf{HyCa}, a training-free framework that reformulates feature caching in diffusion transformers as hybrid ODE solving. It clusters feature dimensions according to their temporal dynamics and assigning tailored solvers to each cluster. Our analysis shows that cluster structures in feature space are input-invariant, enabling \emph{``One-Time Choosing''} and \emph{``All-Time Solving''} with negligible overhead. Extensive experiments demonstrate that HyCa delivers near-lossless acceleration across text-to-image, text-to-video, and image-editing tasks, while remaining compatible with distilled models. HyCa provides a principled and versatile foundation for scaling efficient diffusion generation.